%% file: Plausibility.tex
\pdfoutput=1
\documentclass[11pt]{article}

\usepackage[final]{style/acl}

\usepackage{times}
\usepackage{latexsym}
\usepackage{tabularx,booktabs}

\usepackage[T1]{fontenc}
\usepackage[utf8]{inputenc}

\usepackage{microtype}

\usepackage{inconsolata}

\usepackage{svg}
\usepackage{graphicx}
\usepackage{siunitx}
\usepackage{amsfonts}
\usepackage{amsmath}
\usepackage{booktabs}
\usepackage{multirow}

\usepackage{graphicx}
\usepackage{algorithm}
\usepackage{enumitem}
\usepackage{tablefootnote}
\usepackage{algpseudocode}
\algrenewcommand\algorithmicrequire{\textbf{Input:}}
\algrenewcommand\algorithmicensure{\textbf{Output:}}

\usepackage{caption}
\usepackage{bbold}
\usepackage{cleveref}
\usepackage{subcaption}
\usepackage{soul, color}
\crefname{section}{§}{§§}
\input{style/preamble.tex}

\title{Plausibly Problematic Questions in \\Multiple-Choice Benchmarks for Commonsense Reasoning}

\newcommand{\instA}{\spadesuit}
\newcommand{\instB}{\diamondsuit}
\newcommand{\authorsep}{\hspace*{10mm} \quad}
\newcommand{\institutesep}{\hspace*{5mm} \quad}

\author{
  Shramay Palta$^{\instA}$\authorsep Nishant Balepur$^{\instA}$\authorsep Peter Rankel$^{\instA}$ \\
  \bf Sarah Wiegreffe$^{\instB}$\authorsep Marine Carpuat$^{\instA}$\authorsep Rachel Rudinger$^{\instA}$ \\\\
  $^\spadesuit$University of Maryland, College Park \institutesep $^\diamondsuit$Allen Institute for AI \abr{(AI2)}\\ 
  \texttt{\{spalta,nbalepur,par,marine,rudinger\}@umd.edu} \\ \texttt{wiegreffesarah@gmail.com}
}

\begin{document}
\maketitle
\input{sections/00-abstract}
\input{sections/10-introduction}
\input{sections/20-methodology}
\input{sections/30-results}
\input{sections/40-llm}
\input{sections/50-related}
\input{sections/60-conclusion}
\input{sections/70-limitations}
\input{sections/80-acknowledgements}

\bibliography{bib/anthology, bib/custom}

\appendix
\input{sections/appendix}
\end{document}

%% file: style/preamble.tex
\usepackage{framed}
\usepackage{mdwlist}
\usepackage{siunitx}
\usepackage{latexsym}
\usepackage{colortbl}
\usepackage{xcolor}
\usepackage{nicefrac}
\usepackage{booktabs}
\usepackage{fnpct}
\usepackage{amsfonts}
\usepackage[T1]{fontenc}
\usepackage{bold-extra}
\usepackage{amsmath}
\usepackage{amssymb}
\usepackage{bm}
\usepackage{graphicx}
\usepackage{mathtools}
\usepackage{microtype}
\usepackage{multirow}
\usepackage{multicol}
\usepackage{todonotes}
\usepackage{xpatch}
\usepackage{latexsym,comment}
\usepackage[normalem]{ulem}

\newcommand*{\missingreference}{{\Huge \colorbox{red}{?reference?}}}
\newcommand*{\missingcitation}{{\Huge \colorbox{red}{?citation?}}}

\makeatletter
\xpatchcmd{\@setref}{\bfseries}{\missingreference}{}{}
\def\@citex[#1]#2{\leavevmode
    \let\@citea\@empty
    \@cite{\@for\@citeb:=#2\do
        {\@citea\def\@citea{,\penalty\@m\ }%
            \edef\@citeb{\expandafter\@firstofone\@citeb\@empty}%
            \if@filesw\immediate\write\@auxout{\string\citation{\@citeb}}\fi
            \@ifundefined{b@\@citeb}{\hbox{\reset@font\missingcitation}%
                \G@refundefinedtrue
                \@latex@warning
                {Citation `\@citeb' on page \thepage \space undefined}}%
            {\@cite@ofmt{\csname b@\@citeb\endcsname}}}}{#1}}
\makeatother

\newcommand{\gem}[1]{\mbox{\textsc{gem}}}
\newcommand{\abr}[1]{\textsc{#1}}

\newcommand{\hidetext}[1]{}
\newcommand{\ignore}[1]{}

\newcommand{\specialcell}[2][c]{%
  \begin{tabular}[#1]{@{}c@{}}#2\end{tabular}}

\newcommand{\specialcellleft}[2][l]{%
\begin{tabular}[#1]{@{}l@{}}#2\end{tabular}}

\newcommand{\rachel}[1]{\textcolor{UMDred}{}}
\newcommand{\shramay}[1]{\textcolor{UMDyellow}{}}
\newcommand{\sarah}[1]{\textcolor{green}{}}
\newcommand{\marine}[1]{\textcolor{purple}{}}
\newcommand{\nishant}[1]{\textcolor{lightblue}{}}
\newcommand{\peter}[1]{\textcolor{pink}{}}

\newcommand{\ydata}{$y_{\text{$dataset$}}$}
\newcommand{\yfull}{$y_{\text{$full$}}$}
\newcommand{\yplaus}{$y_{\text{$plausibility$}}$}

\newcommand{\smallurl}[1]{ \begin{tiny}\url{#1}\end{tiny}}

\definecolor{lightblue}{HTML}{3cc7ea}
\definecolor{CUgold}{HTML}{CFB87C}
\definecolor{grey}{rgb}{0.95,0.95,0.95}
\definecolor{ceil}{rgb}{0.57, 0.63, 0.81}
\definecolor{UMDred}{HTML}{ed1c24}
\definecolor{UMDyellow}{HTML}{ffc20e}

%% file: sections/00-abstract.tex
\begin{abstract}
Questions involving commonsense reasoning about everyday situations often admit many \textit{possible} or \textit{plausible} answers. In contrast, multiple-choice question (\abr{MCQ}) benchmarks for commonsense reasoning require a hard selection of a single correct answer, which, in principle, should represent the \textit{most} plausible answer choice. On $250$ \abr{MCQ} items sampled from two commonsense reasoning benchmarks, we collect $5,000$ independent plausibility judgments on answer choices. We find that for over $20\%$ of the sampled \abr{MCQs}, the answer choice rated most plausible does not match the benchmark gold answers; upon manual inspection, we confirm that this subset exhibits higher rates of problems like ambiguity or semantic mismatch between question and answer choices. Experiments with \abr{LLMs} reveal low accuracy and high variation in performance on the subset, suggesting our plausibility criterion may be helpful in identifying more reliable benchmark items for commonsense evaluation\footnote{Our data is available at \url{https://github.com/shramay-palta/commonsense-mcq-plausibility}}.
\end{abstract}

%% file: sections/10-introduction.tex
\section{Introduction}
Commonsense reasoning about everyday situations involves soft judgments about the relative \textit{plausibility} or \textit{likelihood} of different possible outcomes. If a wine glass falls, a \textit{very likely} outcome is that it breaks, but another \textit{technically possible} outcome is that it bounces (e.g., because it lands on a trampoline). Datasets like the Choice of Plausible Alternatives \cite[\abr{COPA};][]{roemmele2011choice} or Ordinal Common-sense Inference \cite{zhang-etal-2017-ordinal} highlight this graded nature of commonsense reasoning.
\begin{figure}[t!]
    \centering    \includegraphics[width=0.90\columnwidth]{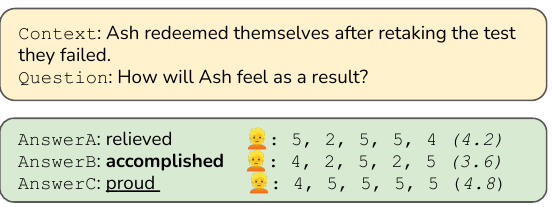}
    \caption{An example question from Social IQa where the highest plausibility answer choice is not the gold label. The numbers indicate the plausibility ratings given by $5$ human annotators to each option on a 1-5 scale and the gold label is highlighted in bold. Numbers in parentheses represent the mean plausibility rating for that answer choice. The answer choice with the highest plausibility rating is underlined.}
    \label{fig:example}
\end{figure}
Many recently developed benchmark datasets for commonsense reasoning formulate problems as multiple choice questions (\abr{MCQs}): \abr{PIQA} \cite{Bisk2020}, Social IQa \cite{sap-etal-2019-social}, CommonsenseQA \cite{talmor-etal-2019-commonsenseqa}, among others. The advantages of \abr{MCQ} evaluation are clear: with a single correct choice per question, system scores are easy to compute and understand. However, by their nature, commonsense reasoning questions typically do not have a single objectively correct answer; rather they admit many possible answers with varying degrees of plausibility as shown in ~\autoref{fig:example}. Under these conditions, what does it mean for a commonsense \abr{MCQ} answer choice to be the ``correct'' answer?

We posit that the ``correct'' \abr{MCQ} answer in this setting should be the one that human annotators agree is \textit{most plausible} among options. In principle, the plausibility of an individual \abr{MCQ} answer choice should depend only on the \abr{MCQ} context (if applicable), question, and the answer choice itself, but need \textit{not} depend on the other answer choices.\footnote{An obvious exception is if an answer choice directly refers to other options, e.g. ``None of the above.''} Under this assumption, then, a valid procedure to determine the correct \abr{MCQ} answer would be to rate the plausibility of each choice individually and select the highest-scoring option. 

In this paper, we analyze two important commonsense \abr{MCQ} benchmarks, Social IQa \cite[\abr{SIQA};][]{sap-etal-2019-social} and CommonsenseQA \cite[\abr{CSQA};][]{talmor-etal-2019-commonsenseqa}, through the lens of this individual plausibility rating procedure. On 250 \abr{MCQ} items sampled from both datasets, we collect 5 Likert-scale plausibility ratings of individual answers in isolation (\cref{plausibility_ratings}), and 5-10 best answer choice judgments given the full set of answers (\cref{full_question_annotations}). With this data, we are able to make the following observations and conclusions:

\begin{enumerate}[nosep]
    \item While gold answers for \abr{MCQs} receive the highest average plausibility rating in a large majority of cases, we observe that, surprisingly, the gold and most-plausible answers do not align in over 20\% of sampled \abr{MCQs} for both datasets.
    \item Through a qualitative analysis of these instances where gold and most-plausible answers do not align, we find a high prevalence of issues such as question ambiguity and answer choices that do not fit the question, among others.
    \item \abr{MCQs} in which the \textit{difference} in mean plausibility scores between the most plausible and second-most plausible answer choices is small are more likely to exhibit low agreement on best answer choice judgments (\cref{results}).
    \item Experiments with LLMs reveal low accuracy and high variation in performance (\cref{LLM_annotations}) on these instances, indicating our approach can help to identify more reliable benchmark items for commonsense evaluation.
\end{enumerate}

%% file: sections/20-methodology.tex
\section{Human Data Collection}
\label{human_annotations}

\begin{table}[t]
\centering
\small
% \resizebox{\columnwidth}{!}{
    \begin{tabular}{c c c c }
    \toprule
    &\textbf{\#MCQ samples} & \textbf{\#Full Anno.} & \textbf{\#Plaus.} \\
    \textbf{Dataset}& \textbf{(\#Answers)} & \textbf{(\#Tie Break)} & \textbf{Ratings} \\
    \midrule
    \abr{SIQA} & $125 (375)$ &$765 (140)$ &$1875$ \\
    \abr{CSQA} & $125 (625)$ &$765 (140)$ &$3125$ \\
    \bottomrule
    Total & $\textbf{250} (1000)$ & $\textbf{1530} (280)$ &$\textbf{5000}$ \\
    \bottomrule
    \end{tabular} 
% }
\caption{Number of annotations performed on Social IQa (\abr{SIQA}) and CommonsenseQA (\abr{CSQA}) samples for the tasks of Individual Plausibility Rating (\cref{plausibility_ratings}) and Full Question Annotation (\cref{full_question_annotations}); totals are bolded.}\label{tab:anno_stats}
\end{table}

We select \abr{CSQA} \cite{talmor-etal-2019-commonsenseqa} and \abr{SIQA} \cite{sap-etal-2019-social} for our study as they are popular \abr{MCQ} benchmarks for general commonsense and social commonsense reasoning, respectively. 

\textbf{Social IQa}: \abr{MCQ} items consist of a short context describing a social situation, a question about a person in the situation, and three answer choices (see Fig.~\ref{fig:example}.) We randomly sample 125 questions from the \texttt{validation} split. These \abr{MCQ} items were originally assigned a gold answer choice based on a majority vote of five annotators. 

\textbf{CommonsenseQA}: \abr{MCQ} items consist of a question generated by humans using \textsc{Conceptnet} \cite{Speer_Chin_Havasi_2017} relations and five possible answer choices. We sample another 125 validation questions, which have gold labels based on approval by a second annotator after construction.

For each of the 250 sampled \abr{MCQ} items from these two datasets, we collect two types of human judgments: \textbf{individual plausibility ratings} (\cref{plausibility_ratings}) and \textbf{full question annotations} (\cref{full_question_annotations}). Annotators are recruited through Prolific and paid \$$15$/hour; see Appendix \ref{annotation_extra} for details, including annotation interfaces (\autoref{ind_sample} and \autoref{full_sample}). Annotation counts for the two tasks are presented in ~\autoref{tab:anno_stats}. 

\subsection{Individual Plausibility Ratings}\label{plausibility_ratings}

\begin{table}[t]
\centering
\resizebox{\columnwidth}{!}{
    \begin{tabular}{c c c}
    \toprule
    \textbf{Statistic}&\textbf{SIQA}&\textbf{CSQA}\\
    \midrule
    Original Gold Answer & $3.86 (0.73)$ & $4.23 (0.71)$ \\
    Maximum Rating & $3.98 (0.67)$ & $4.33 (0.63)$ \\
    Second-Best Rating& $2.88 (0.74)$ & $3.23 (0.99)$ \\
    Minimum Rating & $2.12 (0.67)$ & $1.43 (0.47)$ \\
    Maximum - Second-Best & $1.10 (0.77)$ & $1.10 (0.83)$ \\
    Maximum - Minimum & $1.86 (0.83)$ & $2.90 (0.67)$ \\
    \bottomrule
    \end{tabular} 
}
\caption{Mean Likert-score for gold answers, most-plausible answers, second-most plausible answers, least-plausible answers, and average differences. Numbers in parentheses represent the standard deviation.}\label{tab:individual_rating_stats}
\end{table}

To obtain the plausibility ratings for each option for a given question, we break down each question $q$ with choices $c_1, c_2, ... c_n$ into pairs $(q, c_i)$, where $n=3$ for \abr{SIQA} and $5$ for \abr{CSQA}.

Each $(q, c_i)$ tuple is presented to annotators where they are instructed to ``rate the plausibility of the answer choice for the given question on a $5$-point Likert scale''. We use the plausibility Likert scale introduced by \citet{zhang-etal-2017-ordinal} for ordinal common-sense inference, defined as \textit{\textbf{1-Impossible}, \textbf{2-Technically Possible}, \textbf{3-Plausible}, \textbf{4-Likely} and \textbf{5-Very Likely}.}

\begin{figure*}[t!]
	\centering
	\begin{subfigure}[b]{0.5\linewidth}
        \includegraphics[width=\linewidth]{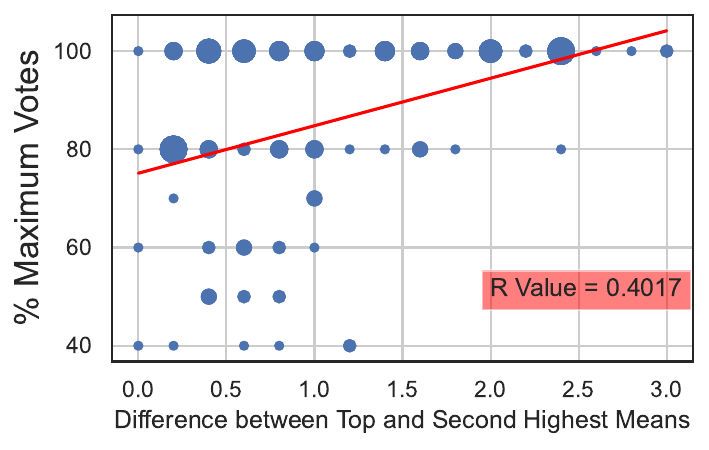}
        \caption{}
        \label{fig:siqa_scatter}
    \end{subfigure}
    \begin{subfigure}[b]{0.49\linewidth}
        \includegraphics[width=\linewidth]{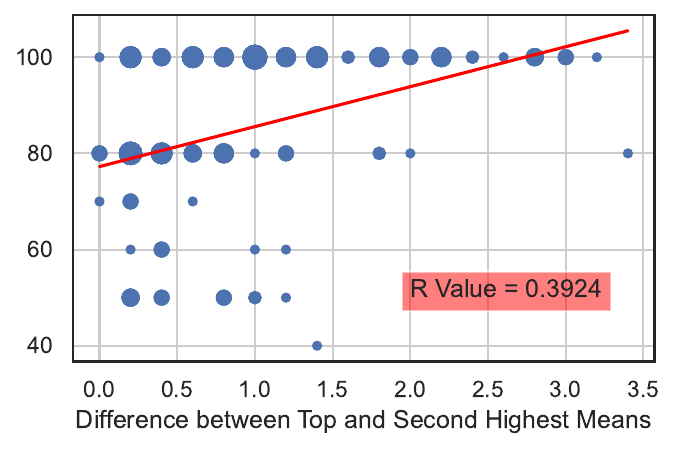}
        \caption{}
        \label{fig:cqa_scatter}
    \end{subfigure}
    \caption{Difference in the plausibility scores between the top 2 most plausible choices (\cref{plausibility_ratings}) vs. percentage of votes (\cref{full_question_annotations}) received by the top choice (on \abr{SIQA} (left) and \abr{CSQA} (right)). The size of the point represents the number of data points at an instance.}
\end{figure*}

We obtain $5$ annotations for each ($q, c_i$) tuple. To ensure independence, each annotator judges at most one $(q, c_i)$ tuple for a given question $q.$ Krippendorff's $\alpha$ on \abr{SIQA} and \abr{CSQA} is $0.46$ and $0.64$, respectively. \footnote{We hypothesise that the low Krippendorff's alpha value for \abr{SIQA} in \cref{plausibility_ratings} is due to the higher difficulty in judging the plausibility of social situations from ATOMIC (\abr{SIQA}) as compared to commonsense knowledge in ConceptNet (\abr{CSQA}).}

For each $(q, c_i)$ tuple, we compute the mean plausibility rating. Mean plausibility statistics are reported in~\autoref{tab:individual_rating_stats}.

\subsection{Full Question Annotation}\label{full_question_annotations}
In this setting, annotators are provided the full \abr{MCQ} item with all the answer choices and asked to select the (single) best option, similar to the validation procedures used to obtain original gold labels. However, to measure human agreement, we re-collect these annotations ourselves in larger numbers. Each \abr{MCQ} item first receives five annotations; if no answer choice receives a majority vote from the annotators by a margin of two or more, then five more annotations are collected for the item. Krippendorff's $\alpha$ on \abr{SIQA} and \abr{CSQA} is $0.66$ and $0.71$, respectively. In over $87\%$ of cases on both datasets, the majority vote from our annotators matches the original gold label in the datasets.

%% file: sections/30-results.tex
\section{Plausibly Problematic \abr{MCQs}}
\label{results}
With these collected judgments, we consider three ways to define a ``correct'' answer choice for each \abr{MCQ} item: (1) the original gold answer choices from \abr{SIQA} or \abr{CSQA} (\ydata), (2) the majority-vote answer choice from full question annotation (\yfull), and (3) the answer choice with the maximum mean plausibility rating (\yplaus). We hypothesize that \yplaus~should be predictive of \ydata~and \yfull~across \abr{MCQs}, and that when they diverge it may be indicative of one or more problems with the underlying \abr{MCQ}.

To corroborate this idea, first we show in~\autoref{fig:siqa_scatter} and~\autoref{fig:cqa_scatter} that a small difference in plausibility scores between the highest- and second-highest scoring answers in the individual plausibility setting is correlated with lower agreement on the full question annotations, for both datasets. \footnote{The p-value for \abr{SIQA} and \abr{CSQA} was evaluated to be $3.42E^{-06}$ and $6E^{-06}$ respectively.} This is consistent with the idea that disagreements on full~\abr{MCQ} annotations may arise when there is not a clear most-plausible answer. 

Next we compare \yplaus~to \ydata. For both \abr{SIQA} and \abr{CSQA}, \yplaus~diverges from \ydata~in $22.4\%$ of \abr{MCQs}. We define these~\abr{MCQs} as ``plausibly problematic'' questions given that the answer choice selected as \yplaus~did not match~\ydata. 

\begin{figure*}[t!]
	\centering
	\begin{subfigure}[b]{0.5\linewidth}
        \includegraphics[width=0.95\columnwidth]{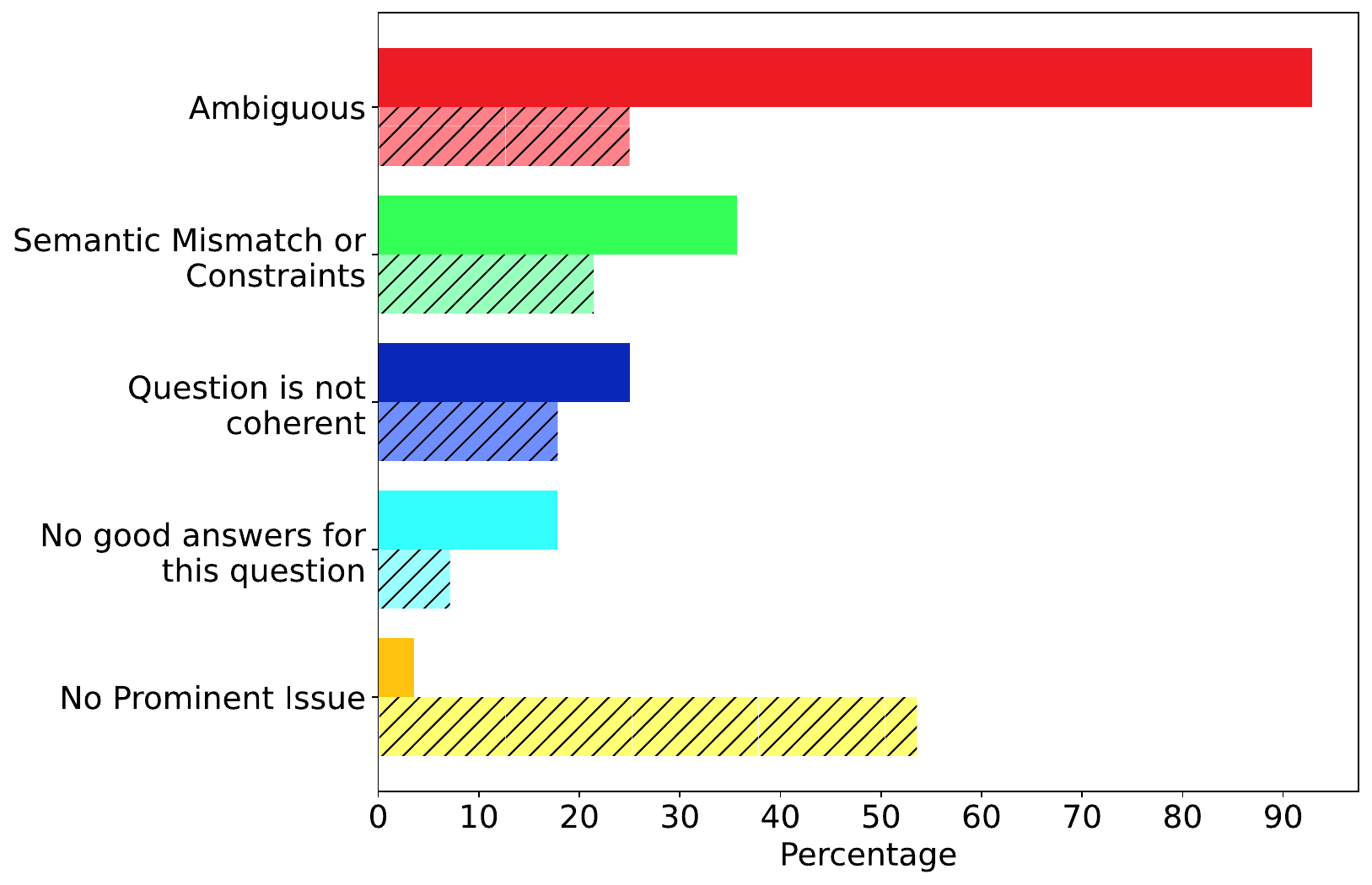}
        \caption{}
        \label{fig:siqa_error_prob}
    \end{subfigure}
    \begin{subfigure}[b]{0.49\linewidth}
        \includegraphics[width=\columnwidth]{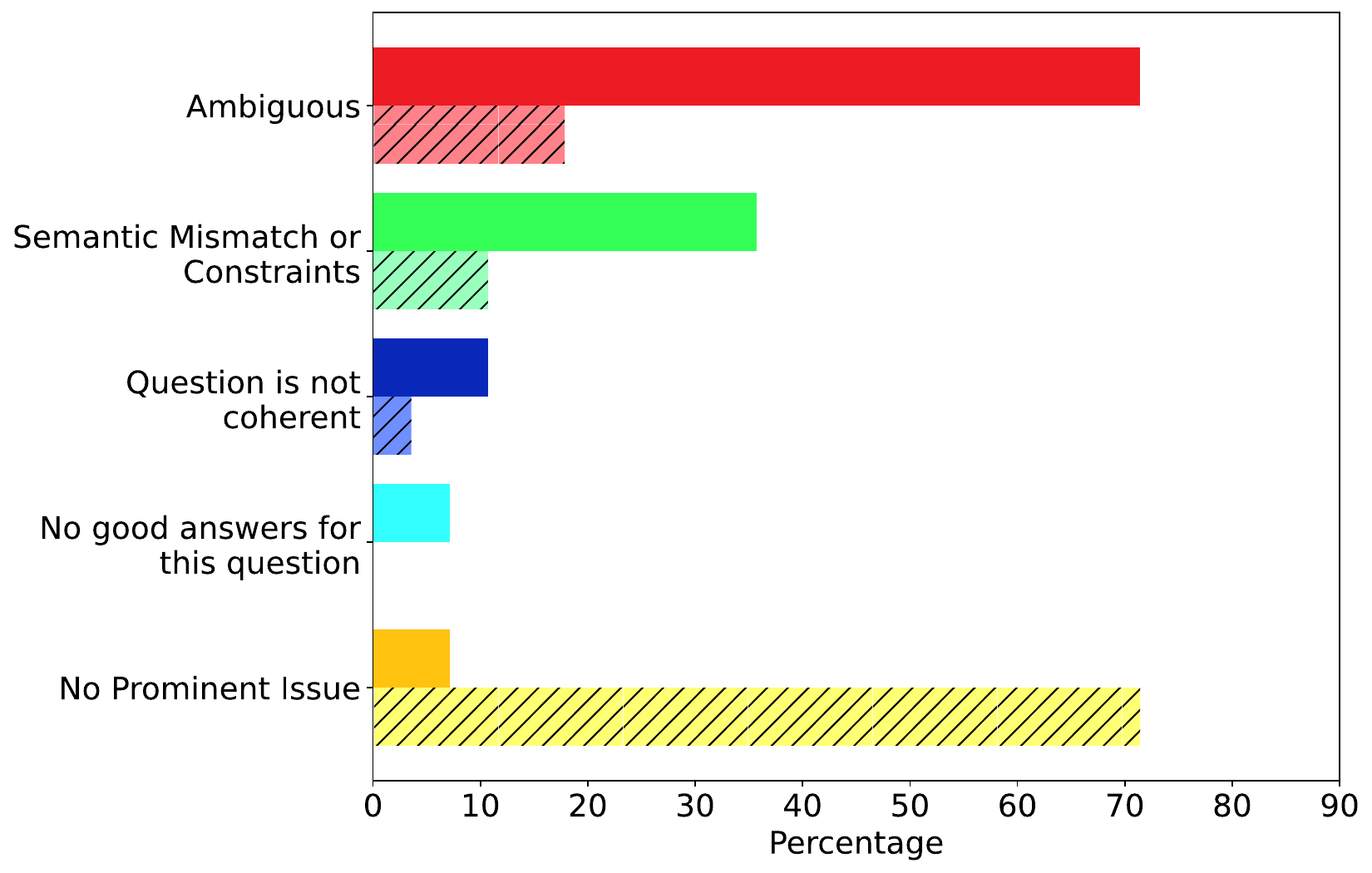}
        \caption{}
        \label{fig:cqa_error_prob}
    \end{subfigure}
    \caption{Frequency of issues types on the ``plausibly problematic'' (solid) and non-problematic (hatched) questions from \abr{SIQA} (left) and \abr{CSQA} (right) ($28$ \abr{MCQs} each). It is important to note that these labels are not mutually exclusive and a question can be ``plausibly problematic'' due to multiple reasons and hence tagged with more than one label.}
\end{figure*}

\subsection{Qualitative Analysis}\label{prob_analysis}
We conduct a manual inspection to identify the key issues with these ``plausibly problematic'' questions (identified using the plausibility judgements from \cref{plausibility_ratings}) and examine all such questions from \abr{SIQA} and \abr{CSQA}. We categorize the potential issues as: \textbf{1)~Semantic Mismatch or Constraints:} A semantic discrepancy exists either between the question and at least one answer choice, or the question implies specific semantic limitations that at least one answer choice fails to meet; \textbf{2)~Question is not coherent:} The question is not properly structured, leading to confusion and lack of clarity or is a poor fit for the context;\footnote{For \abr{SIQA}, we concatenate the context and question as the question, as \abr{CSQA} has no `context' field.} \textbf{3)~Ambiguous:} The question requires one or more implicit assumptions to pick an answer (see \autoref{fig:example}); \textbf{4)~No good answer choices:} There are no answer choices that are a good fit for the question; and \textbf{5)~No Prominent Issue:} There is no prominent issue with the question. Examples of questions with each of these labels are presented in~\autoref{error_examples} in Appendix \ref{error_analysis}.

As seen in \autoref{fig:siqa_error_prob} and \autoref{fig:cqa_error_prob}, \textbf{Ambiguous} and \textbf{Semantic Mismatch or Constraints} are the most common issues with the ``plausibly problematic'' questions. The prevalence of these labels indicates that there are questions in both of these datasets which have multiple possible valid interpretations. We recommend future works to build upon our findings and urge dataset creators to ensure that the questions in their datasets do not have multiple different but valid interpretations, and that all answer choices should be geared towards one interpretation. We also encourage dataset creators to include of ``not applicable'' or ``question does not make sense'' option \cite{dowty1991thematic}, especially when creating datasets involving automatic assignments of questions.

We also observe very few cases where a question is tagged with the \textbf{No Prominent Issue} label, which could be attributed to noise from the human annotations (\cref{plausibility_ratings}). A similar analysis on an equal number of questions sampled randomly from the set of ``Non-Problematic'' Questions is presented in \autoref{fig:siqa_error_prob} and \autoref{fig:cqa_error_prob}. We find that a vast majority of the non-problematic questions would receive the \textbf{No Prominent Issue} label, suggesting that the questions were clear and had an answer choice which was clearly suited better than the others. This indicates that our approach is also able to identify non-problematic questions accurately.

%% file: sections/40-llm.tex
\section{Implications for LLM Evaluation}
\label{LLM_annotations} 
We prompt \abr{LLMs} with the same task posed to humans in \cref{plausibility_ratings} and \cref{full_question_annotations}.
We study multiple state-of-the-art \abr{LLMs}: GPT-4 (\texttt{gpt-4-0125-preview}) \cite{openai2023gpt4} with the OpenAI API, LLaMA-2 (7B, 13B and 70B) \cite{touvron2023llama}, Mistral (7B and 7x8B) \cite{jiang2024mixtral} and Yi (6B, 9B and 34B) \cite{ai2024yi}.
We prompt each \abr{LLM} with the same $10$ in-context examples for the Plausibility and Full settings. \footnote{We present the questions used for in-context examples in ~\autoref{sec:appendix}.} 
\input{data/quant}

We compare human and \abr{LLM} performance on the set of ``plausibly problematic'' and non-problematic questions (identified using the plausibility ratings (\cref{plausibility_ratings})) and present the accuracy (against \ydata) in~\autoref{table:human_llm_perf_ind}. We observe that (1) accuracy on the ``plausibly problematic'' subset is lower, and (2) the performance drop in the problematic set is larger for \abr{LLMs} than for humans. The overall lower performance on the ``plausibly problematic'' subset also suggests that these questions are not merely hard to answer for the models, but have certain underlying issues associated with them, which we discussed in \cref{prob_analysis}.

%% file: data/quant.tex
\begin{table}[ht!]
%\small
\centering
\resizebox{0.9\columnwidth}{!}{
    \begin{tabular}{@{}l|ccc|ccc@{}}
    \toprule
    \multicolumn{1}{c|}{\multirow{2}{*}{\textbf{Agent}}} & \multicolumn{3}{c|}{\textbf{\abr{SIQA}}} & \multicolumn{3}{c}{\textbf{\abr{CSQA}}} \\
    \multicolumn{1}{c|}{} & \multicolumn{1}{l}{Prob} & \multicolumn{1}{l}{Non} & \multicolumn{1}{l|}{All} & \multicolumn{1}{l}{Prob} & \multicolumn{1}{l}{Non} & \multicolumn{1}{l}{All} \\ \midrule
    LLaMA-2 7B & 53.8 & 67.4 & 64.3 & 55.6 & 67.0 & 64.3 \\
    LLaMA-2 13B & 42.3 & 75.3 & 67.8 & 55.6 & 77.3 & 72.2 \\
    LLaMA-2 70B & 57.7 & 87.6 & 80.9 & 66.7 & 85.2 & 80.9 \\
    Mistral 7B & 38.5 & 80.9 & 71.3 & 59.3 & 76.1 & 72.2 \\
    Mixtral 7x8B & 53.8 & 86.5 & 79.1 & 66.7 & 87.5 & 82.6 \\
    Yi 6B & 50.0 & 84.3 & 76.5 & 63.0 & 84.1 & 79.1 \\
    Yi 9B & 73.1 & 91.0 & 87.0 & 74.1 & 85.2 & 82.6 \\
    Yi 34B & 61.5 & 94.4 & 87.0 & 70.4 & 90.9 & 86.1 \\
    GPT-4 & 53.8 & 89.9 & 81.7 & 59.3 & 92.0 & 84.3 \\\midrule
    Average \abr{LLM} & 53.8 & 84.1 & 77.3 & 63.4 & 82.8 & 78.3 \\\midrule
    Human & 71.2 & 94.4 & 89.1 & 70.4 & 92.6 & 87.4 \\ \bottomrule
    \end{tabular}
    }
\caption{\label{table:human_llm_perf_ind}Percentages of cases where the agent response in the full question setting matches the original dataset gold label on the set of ``plausibly problematic'' and non-problematic questions (identified using plausibility judgements from \cref{plausibility_ratings}) from \abr{SIQA} and \abr{CSQA}.}
\end{table}

%% file: sections/50-related.tex
\section{Related Works}
\paragraph{Dataset Quality Analysis:} Many works find biases in datasets, including dataset artifacts \cite{poliak-etal-2018-hypothesis, gururangan-etal-2018-annotation, balepur-etal-2024-artifacts, balepur-rudinger-2024-large} and annotator noise~\cite{sheng2008, snow-etal-2008-cheap, oro25874}. Given these findings, recent work has proposed not to treat every data entry as equally difficult when assessing LMs \cite{rodriguez-etal-2021-evaluation}, using human psychology techniques such as Item Response Theory \cite{lalor-etal-2016-building, vania-etal-2021-comparing, rodriguez-etal-2022-clustering} or model-based hardness metrics \cite{perez2021rissanen}. \citet{swayamdipta-etal-2020-dataset} use this method to disentangle difficult and ambiguous/noisy data entries. Similarly, we show how plausibility ratings can uncover problematic data in \abr{MCQ} datasets.

\paragraph{Plausibility in Commonsense:} Ranking, comparing, and scoring the plausibility of events and outcomes expressed in language is a long-standing concept in commonsense reasoning research\cite{roemmele2011choice,wang2018modeling,li-etal-2019-learning,liu-etal-2023-vera}. Because commonsense knowledge is often subjective \cite{whiting2024framework} or graded \cite{zhang-etal-2017-ordinal,chen-etal-2020-uncertain}, and varies with cultural context \cite{palta-rudinger-2023-fork, hershcovich-etal-2022-challenges, bhatia-shwartz-2023-gd}, this can pose challenges for evaluation.  Most relevant to this work, \citet{acquaye-an-rudinger-2024-susu} use Likert-scale human plausibility judgments of answer choices to construct cultural commonsense \abr{MCQ} test items. Other approaches to evaluation include verbalized rationales \cite{jung2022maieutic, balepur-etal-2024-easy}. Specifically, prior works have studied defeasible \cite{rudinger-etal-2020-thinking, rao-etal-2023-makes} and abductive reasoning \cite{bhagavatula2020abductive} in natural language, where models rationalize when scenarios may be more plausible or valid.

%% file: sections/60-conclusion.tex
\section{Conclusion}
In this work, we show that plausibility judgments are a useful tool for identifying \textit{potentially} problematic commonsense \abr{MCQ} items. With individual plausibility ratings, we are able to identify questions where the gold answer does not match the answer with the highest plausibility. Through manual analysis we identify several types of issues that are more prevalent among the identified subset. We show that \abr{LLMs} and humans perform poorly on these questions, with a high degree of variance, suggesting they add noise to benchmark evaluations. Future work may investigate methods of incorporating plausibility judgments into the creation stage of benchmark development, as well as the application of these ideas to evaluating other types of benchmarks involving graded judgments beyond commonsense reasoning.

%% file: sections/70-limitations.tex
\section{Limitations}
Uncertainty can arise due to a variety of reasons such as multi-cultural and multi-ethnic aspects of commonsense reasoning. In this work while we introduce a new method to identify questions with multiple plausible answers, we are limited to a US-centric angle of uncertainty owing to the fact that our annotators are based in the US. 

Additionally, our annotation framework is expensive and thus difficult to run on an entire dataset. However, since we are the first to explore plausibility of answer choices in commonsense reasoning situations, we hope that this work motivates other researchers to study plausibility more extensively.

The identification and annotation of uncertainty can be subjective, leading to inconsistencies or disagreements among annotators. While we employed rigorous annotation protocols and made sure each question was annotated by at least $5$ annotators, there may still be instances where ambiguity interpretation varies.

%% file: sections/80-acknowledgements.tex
\section{Acknowledgements}
We would like to thank the anonymous reviewers for their valuable feedback on this paper. We would also like to thank Hal Daumé III, Faeze Brahman, Elijah Rippeth, Alexander Hoyle, Yuelin Liu, Sander Schulhoff, Abhilasha Sancheti, Haozhe An, Neha Srikanth, Christabel Acquaye, Yu Hou and other members of the \abr{CLIP} lab for for their helpful comments and suggestions. Nishant Balepur's funding and this material is based upon work supported by the National Science Foundation Graduate Research Fellowship Program under Grant No. DGE 2236417. Rachel Rudinger is supported by NSF CAREER Award No.~2339746. Any opinions, findings, and conclusions or recommendations expressed in this material are those of the author(s) and do not necessarily reflect the views of the National Science Foundation.

%% file: sections/appendix.tex
\section{Appendix}
\label{sec:appendix}

\subsection{License for Artifacts}
All datasets used in this work are publicly available and free to use on HuggingFace.

\subsection{Details on Computational Experiments}

LLaMA-2 70B, LLaMA-2-13B, Yi-34B, Yi-9B, and Mixtral 7x8B were all run on eight NVIDIA:RTXA5000 GPUs and were allocated a total of eight GPU hours to run all experiments. All other open-source LLMs were run on one NVIDIA:RTXA6000 GPU and were allocated a total of two GPU hours to run all experiments. GPT-4 was run on CPU and was allocated one hour to run all experiments. Each LLM decodes with a minimum token generation length of 5, a maximum token generation length of 200, greedy decoding (or 0 temperature in the case of GPT-4), and a stopping criteria when the LLM begins to generate the next few-shot exemplar. We did not perform a hyperparameter search. All results are obtained from a single run. 

\subsection{Additional Experiments and Results}
\label{extra_data}

\subsubsection{Full Question Setting}
\label{full_ques_exp}
In this setting, for humans, we look at the vote distribution for each question and use that to determine whether \yfull $=$ \ydata. We flag the questions as ``problematic'' in the \textit{Full Question Setting} if \yfull~$\neq$ \ydata~or the difference between the highest and second highest votes (for humans) is less than $2$.

We observe that in the \textit{Full Question setting}, humans exhibit overall better performance than \abr{LLMs} (highlighted in \autoref{table:human_llm_perf_full}), suggesting that even when presented with all the answer choices, 'problematic' questions pose a challenge to effective \abr{LLM} evaluation of commonsense reasoning capabilities.

\input{data/full_ind_consistency}

\autoref{tab:individual_to_full} demonstrates the cases where \yplaus $=$ \yfull. \autoref{tab:pearson_corr} shows the Pearson's Correlation Coefficient for Human and \abr{LLM} individual plausibility ratings.

\autoref{table:human_llm_match_ind} and \autoref{table:human_llm_match_full} demonstrate that \abr{LLMs} show higher agreement with the human responses in cases where the questions are not identified as problematic. This finding is consistent in both the \textit{Individual Plausibility Setting} (\cref{plausibility_ratings}) and the \textit{Full Question Setting} (\cref{full_question_annotations}).

\begin{table}[t]
\small
\centering
\begin{tabular}{ c c c }
\toprule
\textbf{Model} & \textbf{\abr{SIQA}} &  \textbf{\abr{CSQA}} \\
\midrule
LLaMA-2 7B & 0.262 & 0.178 \\
LLaMA-2 13B & 0.417 & 0.654 \\
LLaMA-2 70B & 0.656 & 0.760 \\
Mistral 7B & 0.385 & 0.675 \\
Mixtral 7x8B & 0.573 & 0.709 \\
Yi 6B & 0.330 & 0.399 \\
Yi 9B & 0.652 & 0.700 \\
Yi 34B & 0.648 & 0.716 \\
GPT-4 & 0.708 & 0.775 \\
\bottomrule
\end{tabular}
\caption{Pearson's Correlation coefficients between \abr{LLM} plausibility ratings and human plausibility ratings.}\label{tab:pearson_corr}
\end{table}

\begin{figure*}[t!]
    \centering
    \begin{subfigure}[b]{0.475\linewidth}
        \centering
        \includegraphics[width=1.02\linewidth]{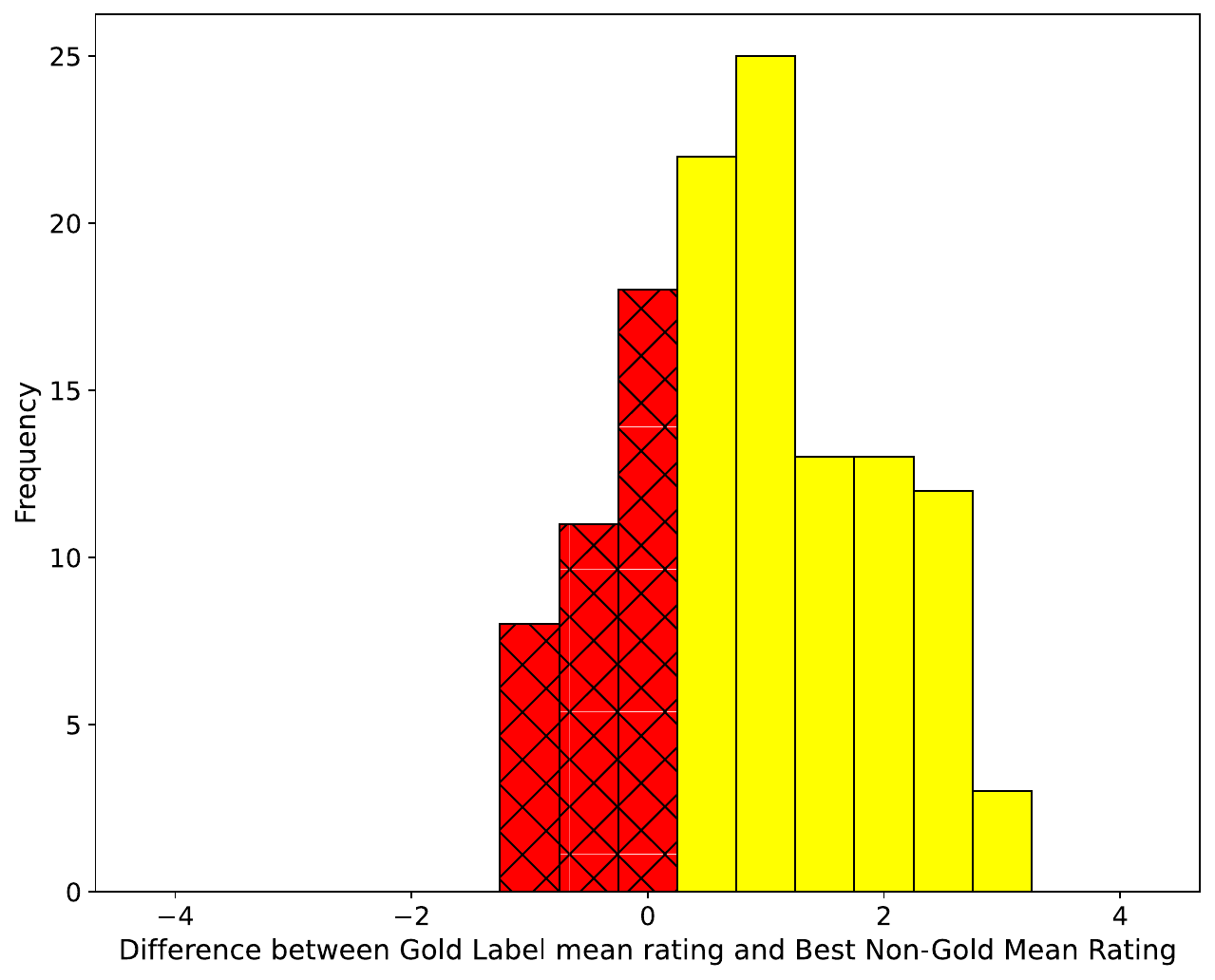}
        \caption{Social IQa}
        \label{fig:siqa_mean_hist}
    \end{subfigure}
    \hfill
    \begin{subfigure}[b]{0.515\linewidth}
        \centering
        \includegraphics[width=0.96\linewidth]{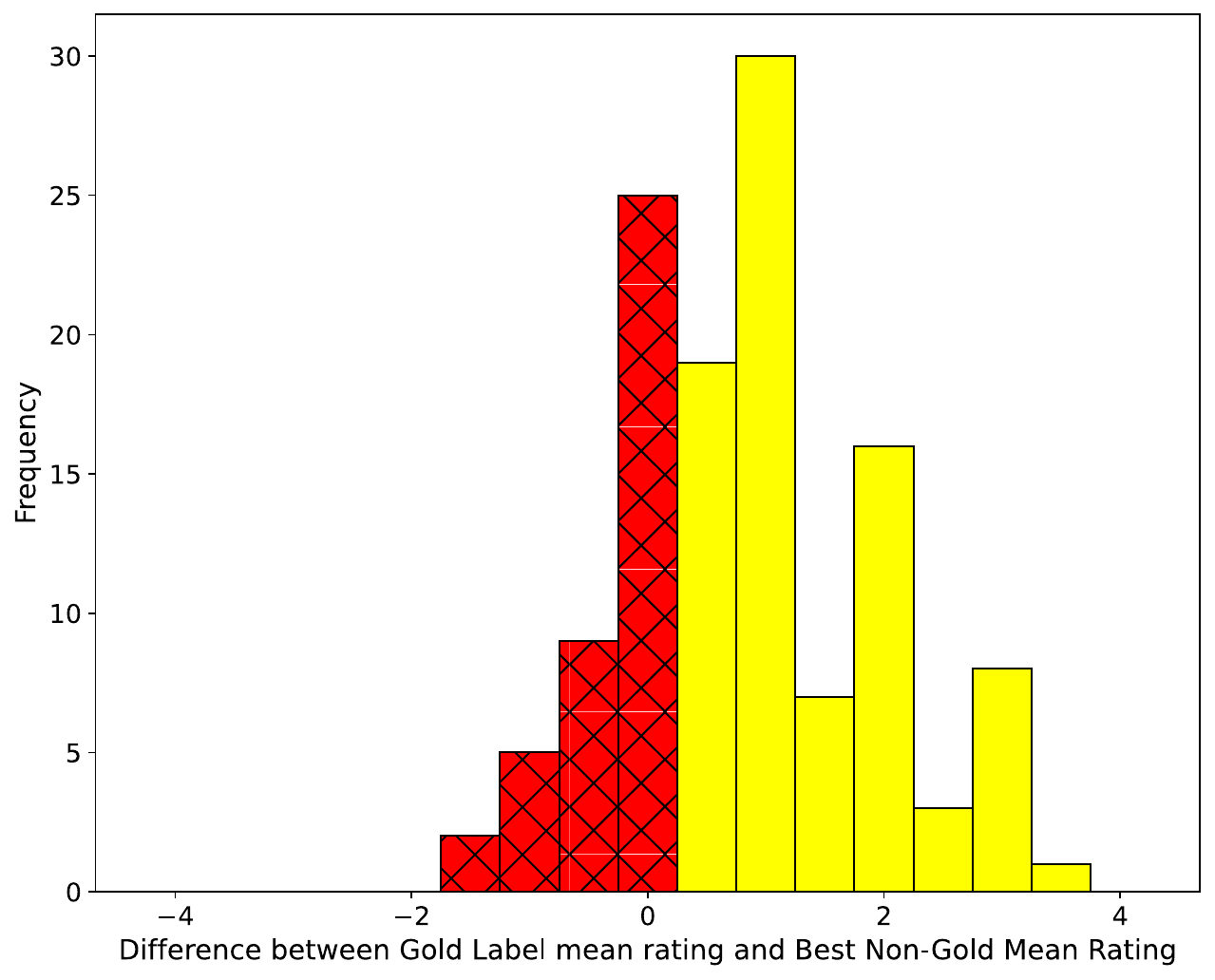}
        \caption{CommonsenseQA}
        \label{fig:cqa_mean_hist}
    \end{subfigure}
    \caption{Histograms showing the difference between the mean gold label rating and best non-gold label rating. Portions of the graph in \textcolor{red}{red} with texture show cases where the best non-gold option had a higher mean plausibility rating than the mean gold label rating.}
\end{figure*}

\begin{table*}[ht!]
\small
\centering
\begin{tabular}{@{}l|ccc|ccc@{}}
\toprule
\multicolumn{1}{c|}{\multirow{2}{*}{\textbf{Model}}} & \multicolumn{3}{c|}{\textbf{\abr{SIQA}}} & \multicolumn{3}{c}{\textbf{\abr{CSQA}}} \\
\multicolumn{1}{c|}{} & \multicolumn{1}{l}{Problematic} & \multicolumn{1}{l}{Non-Problematic} & \multicolumn{1}{l|}{Overall} & \multicolumn{1}{l}{Problematic} & \multicolumn{1}{l}{Non-Problematic} & \multicolumn{1}{l}{Overall} \\ \midrule
LLaMA-2 7B & 61.5 & 67.4 & 66.1 & 70.4 & 68.2 & 68.7 \\
LLaMA-2 13B & 51.9 & 72.5 & 67.8 & 63.0 & 75.0 & 72.2 \\
LLaMA-2 70B & 50.0 & 87.6 & 79.1 & 81.5 & 83.5 & 83.0 \\
Mistral 7B & 48.1 & 78.7 & 71.7 & 74.1 & 77.8 & 77.0 \\
Mixtral 7x8B & 51.9 & 84.3 & 77.0 & 77.8 & 85.8 & 83.9 \\
Yi 6B & 53.8 & 82.6 & 76.1 & 66.7 & 79.0 & 76.1 \\
Yi 9B & 59.6 & 89.9 & 83.0 & 77.8 & 83.5 & 82.2 \\
Yi 34B & 55.8 & 94.4 & 85.7 & 77.8 & 91.5 & 88.3 \\
GPT-4 & 59.6 & 88.8 & 82.2 & 74.1 & 88.1 & 84.8 \\ \bottomrule
\end{tabular}
\caption{\label{table:human_llm_match_ind} Instances where the \abr{LLM} response matches the response given by humans, based on  maximum vote on the set of ``plausibly problematic'' and non-problematic questions (identified from the \textit{Individual Plausibility Rating Setting}) in the Full Question Setting.}
\end{table*}

\begin{table*}[ht!]
\small
\centering
\begin{tabular}{@{}l|ccc|ccc@{}}
\toprule
\multicolumn{1}{c|}{\multirow{2}{*}{\textbf{Agent}}} & \multicolumn{3}{c|}{\textbf{\abr{SIQA}}} & \multicolumn{3}{c}{\textbf{\abr{CSQA}}} \\
\multicolumn{1}{c|}{} & \multicolumn{1}{l}{Problematic} & \multicolumn{1}{l}{Non-Problematic} & \multicolumn{1}{l|}{Overall} & \multicolumn{1}{l}{Problematic} & \multicolumn{1}{l}{Non-Problematic} & \multicolumn{1}{l}{Overall} \\ \midrule
LLaMA-2 7B & 45.5 & 66.3 & 62.9 & 33.3 & 66.1 & 62.1 \\
LLaMA-2 13B & 45.5 & 70.2 & 66.1 & 33.3 & 74.3 & 69.3 \\
LLaMA-2 70B & 45.5 & 84.6 & 78.1 & 33.3 & 83.5 & 77.4 \\
Mistral 7B & 45.5 & 74.0 & 69.3 & 33.3 & 74.3 & 69.3 \\
Mixtral 7x8B & 45.5 & 82.7 & 76.6 & 50.0 & 84.4 & 80.2 \\
Yi 6B & 45.5 & 79.8 & 74.1 & 66.7 & 79.8 & 78.2 \\
Yi 9B & 72.7 & 88.5 & 85.9 & 50.0 & 84.4 & 80.2 \\
Yi 34B & 81.8 & 87.5 & 86.6 & 50.0 & 88.1 & 83.5 \\
GPT-4 & 54.5 & 84.6 & 79.6 & 50.0 & 86.2 & 81.8 \\\midrule
Average \abr{LLM} & 53.6 & 79.8 & 75.5 & 44.4 & 80.1 & 75.8 \\\midrule
Human & 22.7 & 96.2 & 84.1 & 8.3 & 91.7 & 81.5 \\ \bottomrule
\end{tabular}
\caption{\label{table:human_llm_perf_full} Percentages of cases where agent response to the full question matches the original dataset gold label on the set of problematic and non-problematic questions (identified from the \textit{Full Question setting}) from \abr{SIQA} and \abr{CSQA}.}
\end{table*}

\begin{table*}[ht!]
\small
\centering
\begin{tabular}{@{}l|ccc|ccc@{}}
\toprule
\multicolumn{1}{c|}{\multirow{2}{*}{\textbf{Model}}} & \multicolumn{3}{c|}{\textbf{\abr{SIQA}}} & \multicolumn{3}{c}{\textbf{\abr{CSQA}}} \\
\multicolumn{1}{c|}{} & \multicolumn{1}{l}{Problematic} & \multicolumn{1}{l}{Non-Problematic} & \multicolumn{1}{l|}{Overall} & \multicolumn{1}{l}{Problematic} & \multicolumn{1}{l}{Non-Problematic} & \multicolumn{1}{l}{Overall} \\ \midrule
LLaMA-2 7B & 27.3 & 70.2 & 63.1 & 33.3 & 70.6 & 66.1 \\
LLaMA-2 13B & 27.3 & 72.1 & 64.7 & 50.0 & 73.4 & 70.6 \\
LLaMA-2 70B & 27.3 & 84.6 & 75.1 & 58.3 & 84.4 & 81.2 \\
Mistral 7B & 31.8 & 76.0 & 68.7 & 58.3 & 78.0 & 75.6 \\
Mixtral 7x8B & 22.7 & 82.7 & 72.8 & 41.7 & 86.2 & 80.8 \\
Yi 6B & 22.7 & 81.7 & 72.0 & 25.0 & 78.9 & 72.3 \\
Yi 9B & 31.8 & 88.5 & 79.1 & 41.7 & 84.4 & 79.2 \\
Yi 34B & 31.8 & 91.3 & 81.5 & 41.7 & 90.8 & 84.8 \\
GPT-4 & 40.9 & 86.5 & 79.0 & 41.7 & 87.2 & 81.6 \\ \bottomrule
\end{tabular}
\caption{\label{table:human_llm_match_full} Instances where the \abr{LLM} response matches the response given by humans, based on  maximum vote on the set of problematic and non-problematic questions (identified from the \textit{Full Question Setting}) in the Full Question Setting.}
\end{table*}

\subsection{Annotation Process Details}\label{annotation_extra}
We used Prolific to collect the human annotations. The annotators for our task were selected on the basis of the following criteria:
\begin{enumerate}[nosep]
    \item Must be located in the United States.
    \item Primary language must be English.
    \item Must not have any literacy difficulties.
    \item Must have attained a minimum of an undergraduate level degree.
    \item Must have an approval rate between $95-100\%$ on Prolific.
    \item We use a $50-50$ split of male and female\footnote{Gender as indicated on Prolific.} annotators to minimize the risk of any gender-specific biases creeping in.
\end{enumerate}

The total cost for our entire annotation protocols, for both Individual Plausibility Ratings, and the Full Question Setting came out to be $\$1052.$ We also received an exempt status from the \abr{IRB} at our institution for this research.

\subsection{Examples of Questions with Labels}\label{error_analysis}
We include an example question for each label used in our error analysis as described in \cref{prob_analysis} and present them in \autoref{error_examples}.
\input{data/error}

\clearpage

\begin{figure*}
    \centering
    \fbox{\includegraphics[width=1\linewidth]{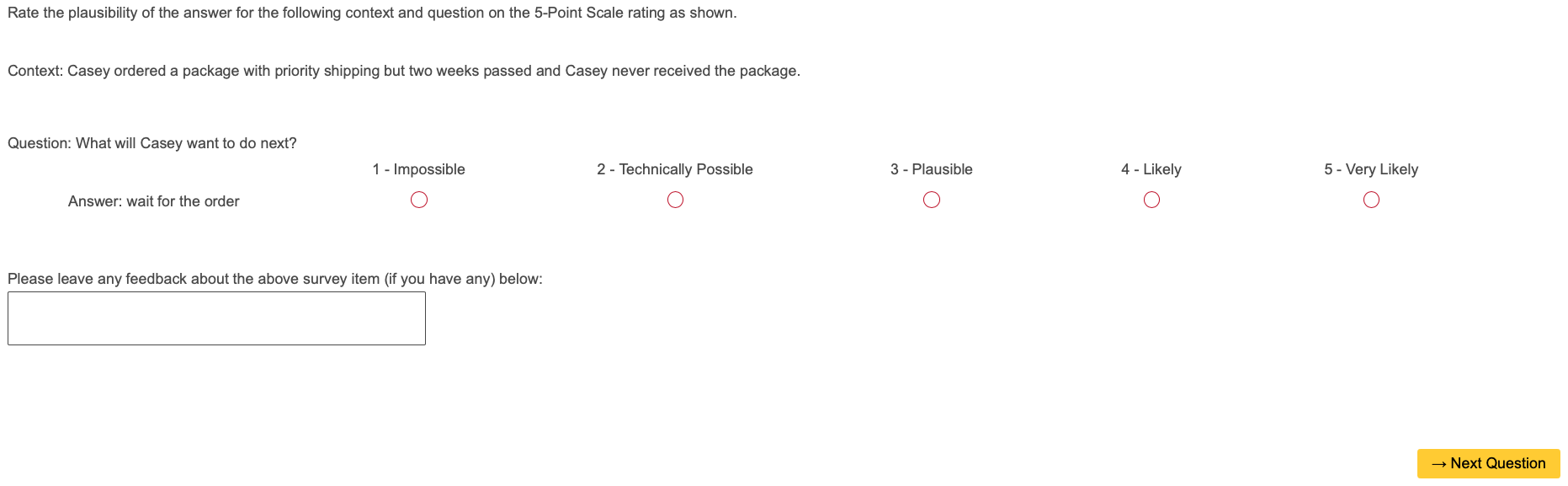}}
    \caption{\label{ind_sample} An example of the interface that annotators used while giving plausibility ratings to answer choices as described in \cref{plausibility_ratings}.}
\end{figure*}

\begin{figure*}
    \centering
    \fbox{\includegraphics[width=1\linewidth]{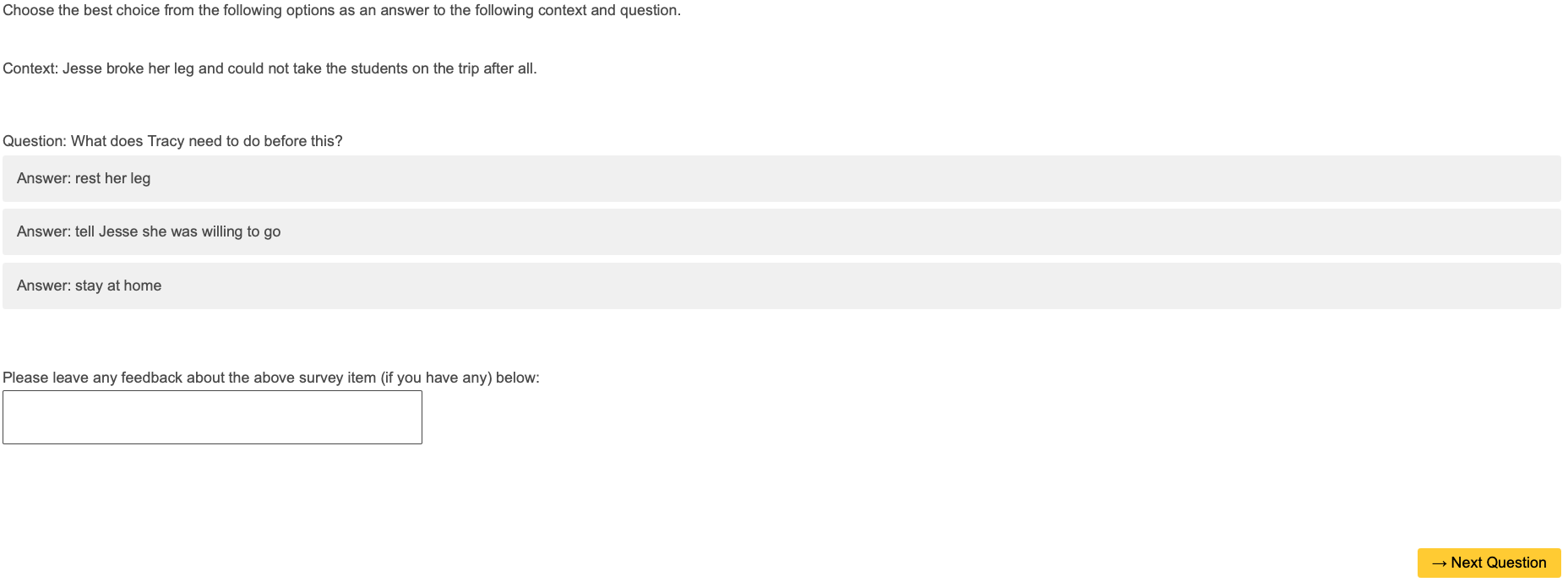}}
    \caption{\label{full_sample} An example of the interface that annotators used while choosing the best answer choice for a question as described in \cref{full_question_annotations}.}
\end{figure*}

\begin{figure*}
    \centering
    \fbox{\includegraphics[width=0.75\linewidth]{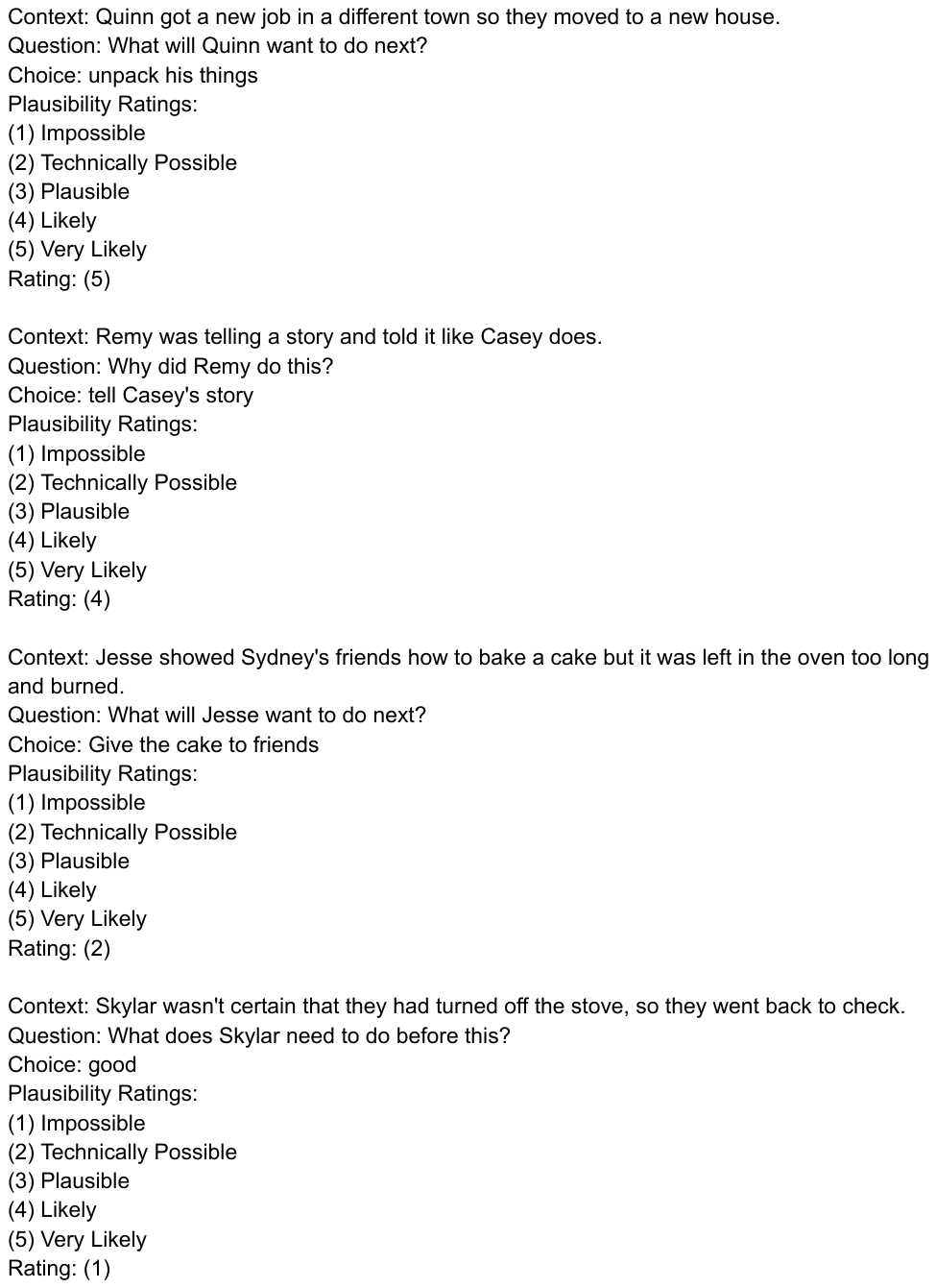}}
    \caption{In-context learning examples from Social IQa for the isolated setting. (Part 1)}
\end{figure*}

\begin{figure*}
    \centering
    \fbox{\includegraphics[width=0.75\linewidth]{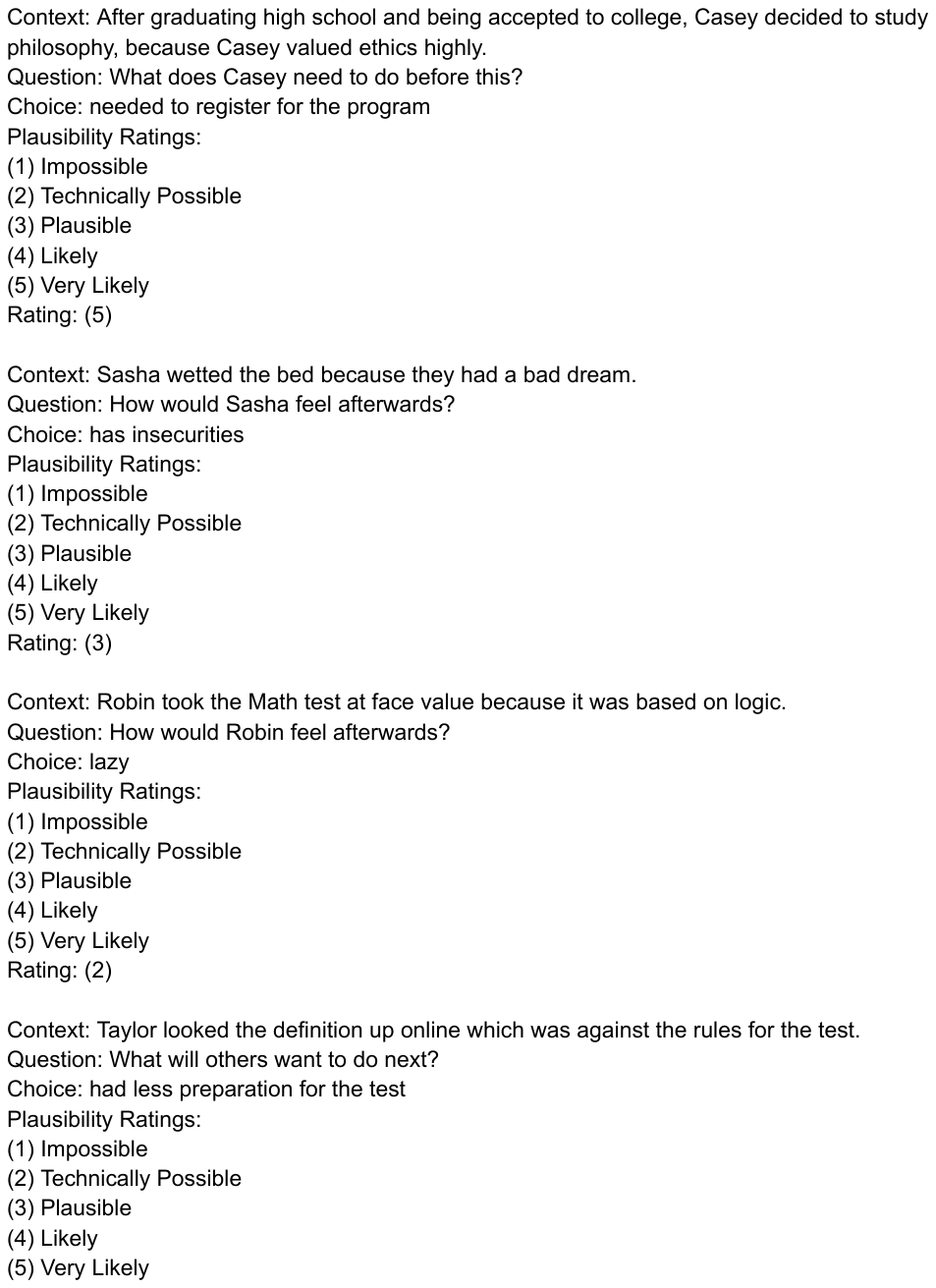}}
    \caption{In-context learning examples from Social IQa for the isolated setting. (Part 2)}
\end{figure*}

\begin{figure*}
    \centering
    \fbox{\includegraphics[width=0.75\linewidth]{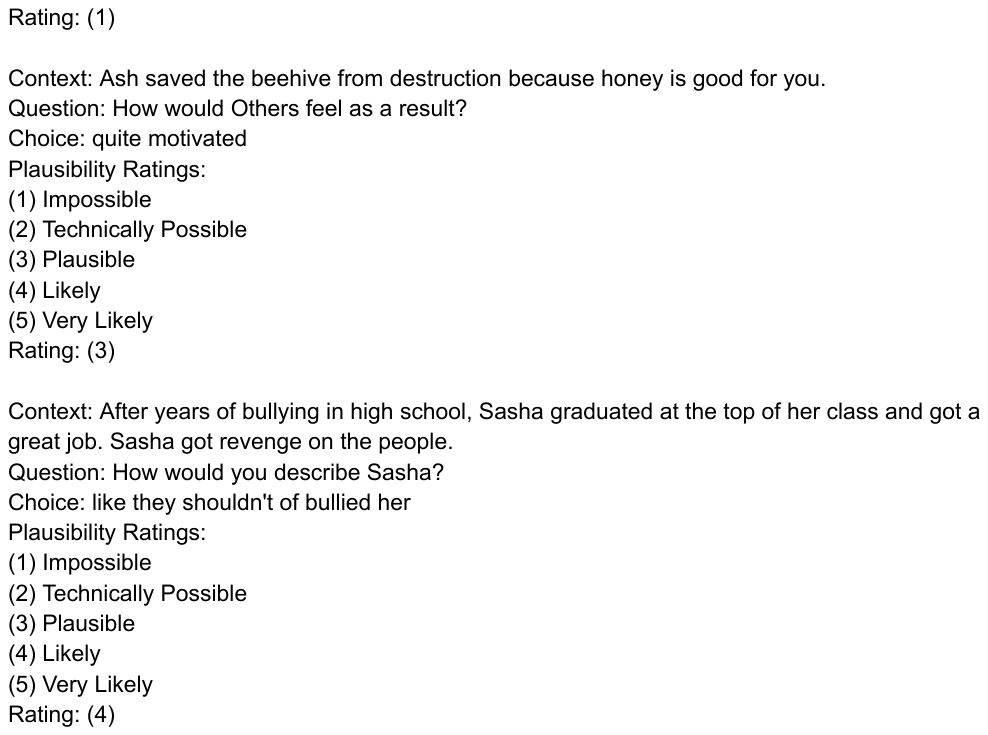}}
    \caption{In-context learning examples from Social IQa for the isolated setting. (Part 3)}
\end{figure*}

\begin{figure*}
    \centering
    \fbox{\includegraphics[width=0.75\linewidth]{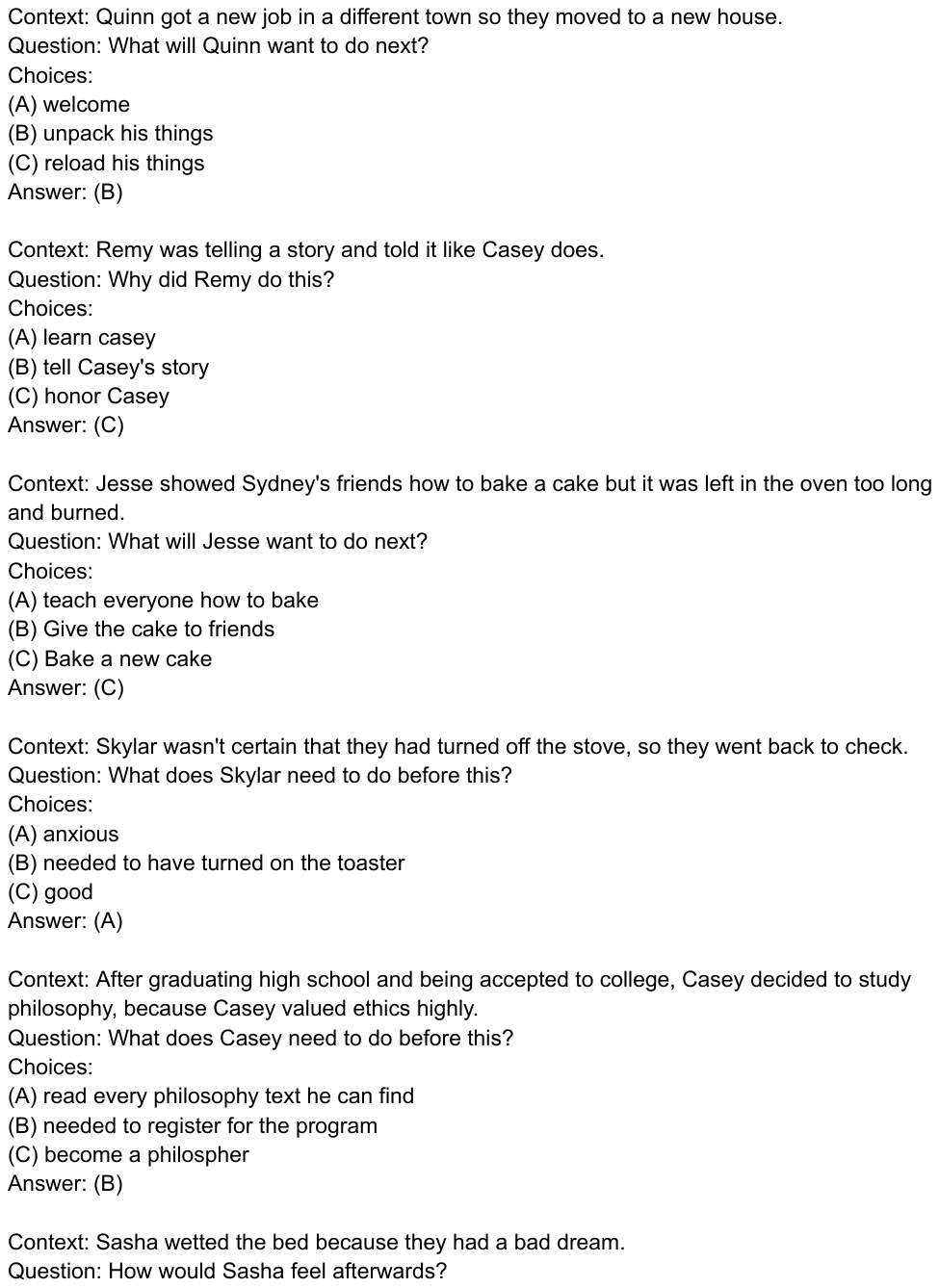}}
    \caption{In-context learning examples from Social IQa for the full setting. (Part 1)}
\end{figure*}

\begin{figure*}
    \centering
    \fbox{\includegraphics[width=0.75\linewidth]{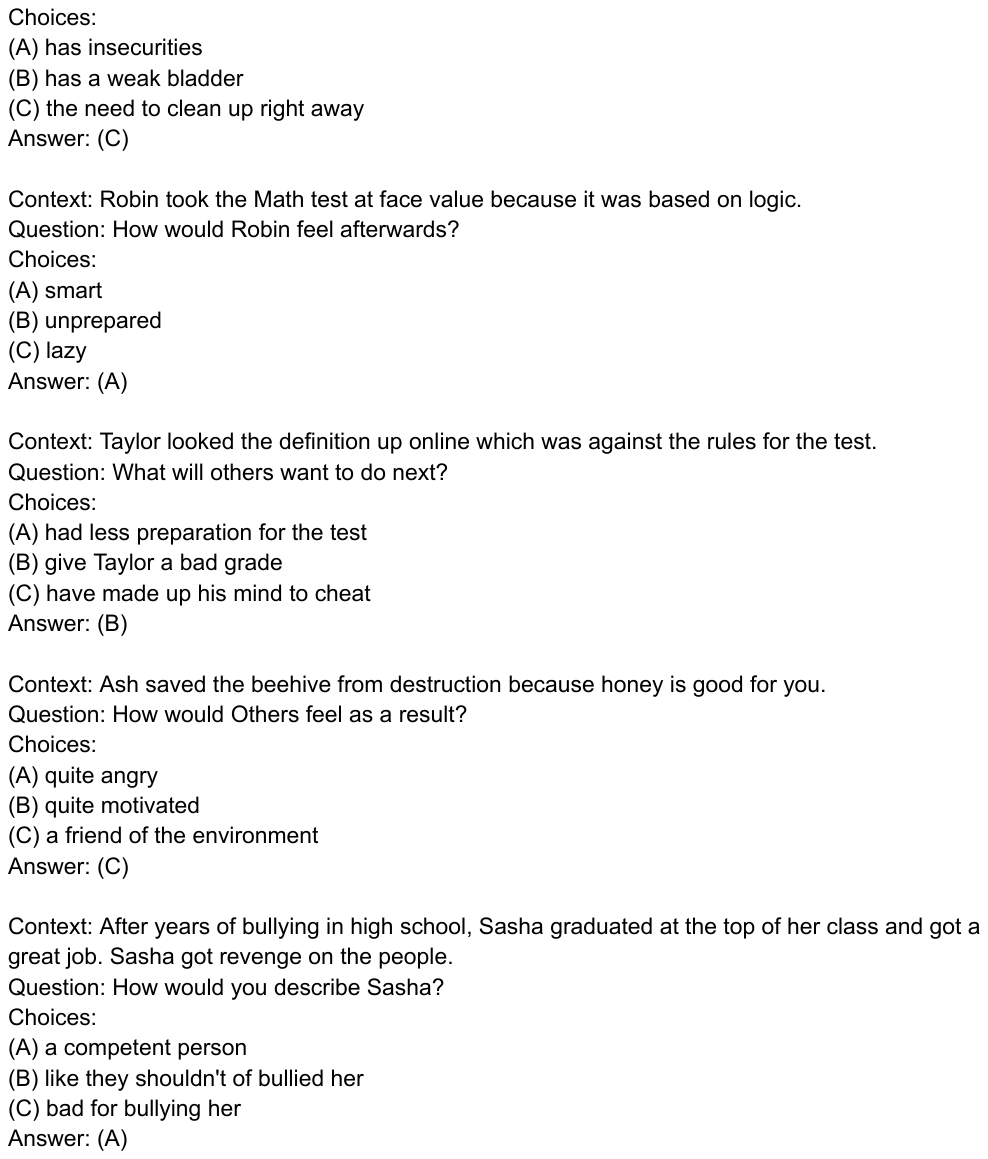}}
    \caption{In-context learning examples from Social IQa for the full setting. (Part 2)}
\end{figure*}

\begin{figure*}
    \centering
    \fbox{\includegraphics[width=0.75\linewidth]{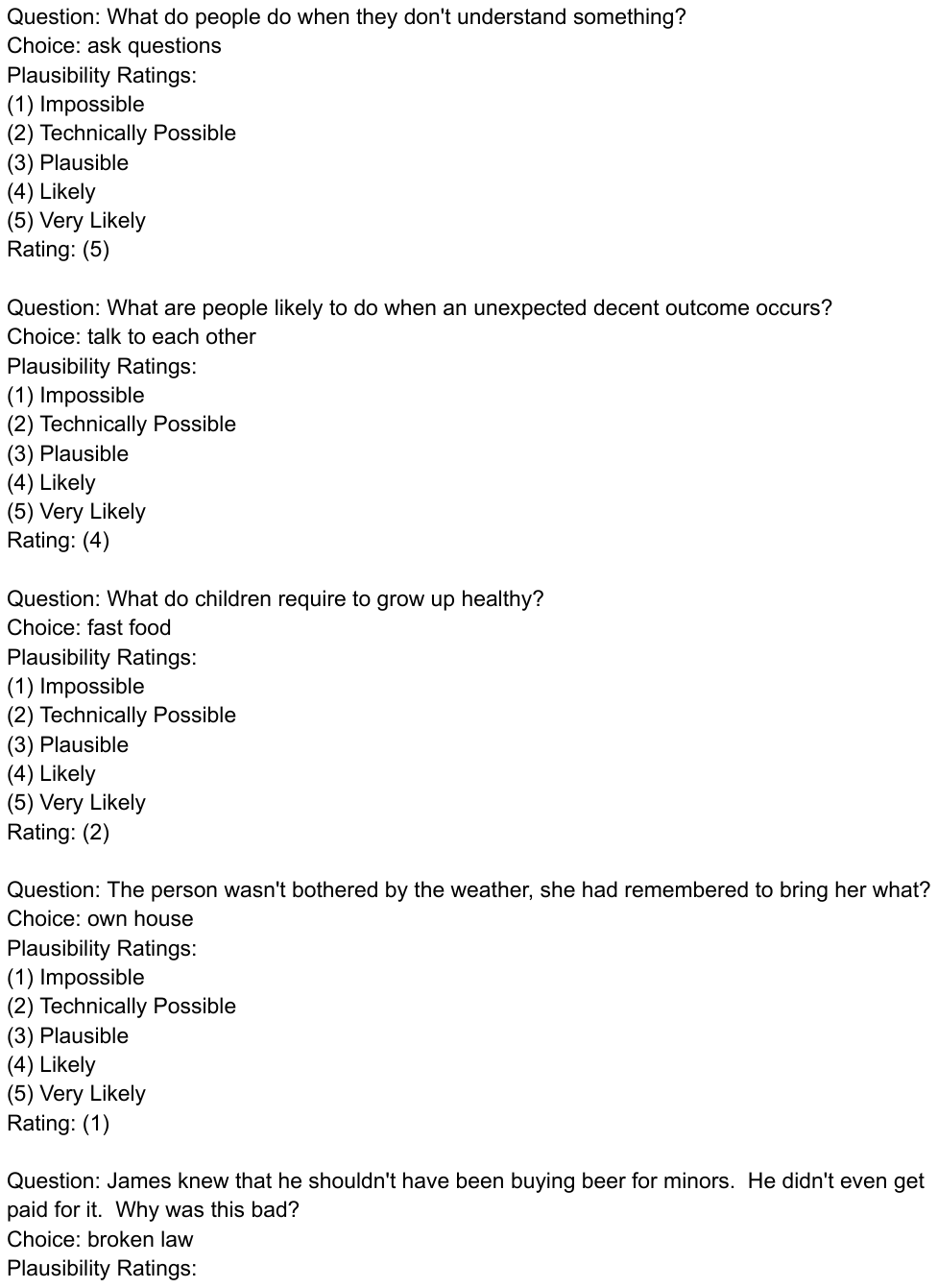}}
    \caption{In-context learning examples from CommonsenseQA for the isolated setting. (Part 1)}
\end{figure*}

\begin{figure*}
    \centering
    \fbox{\includegraphics[width=0.75\linewidth]{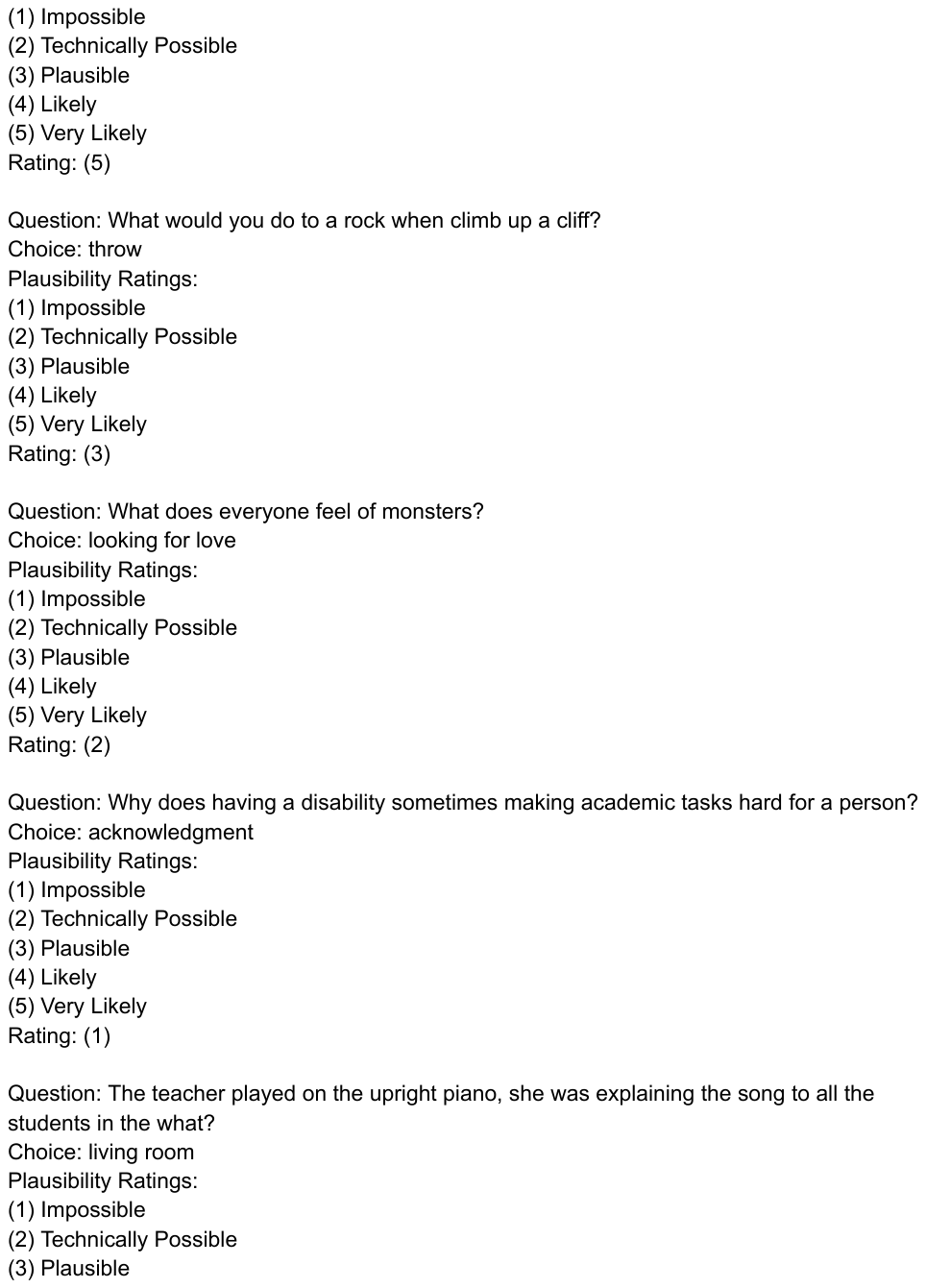}}
    \caption{In-context learning examples from CommonsenseQA for the isolated setting. (Part 2)}
\end{figure*}

\begin{figure*}
    \centering
    \fbox{\includegraphics[width=0.75\linewidth]{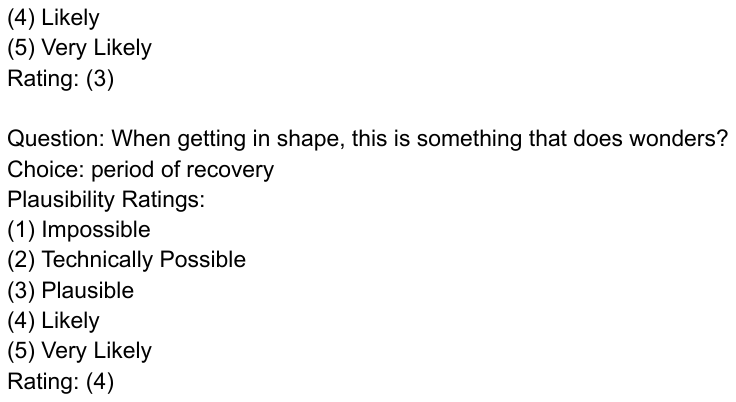}}
    \caption{In-context learning examples from CommonsenseQA for the isolated setting. (Part 3)}
\end{figure*}

\begin{figure*}
    \centering
    \fbox{\includegraphics[width=0.75\linewidth]{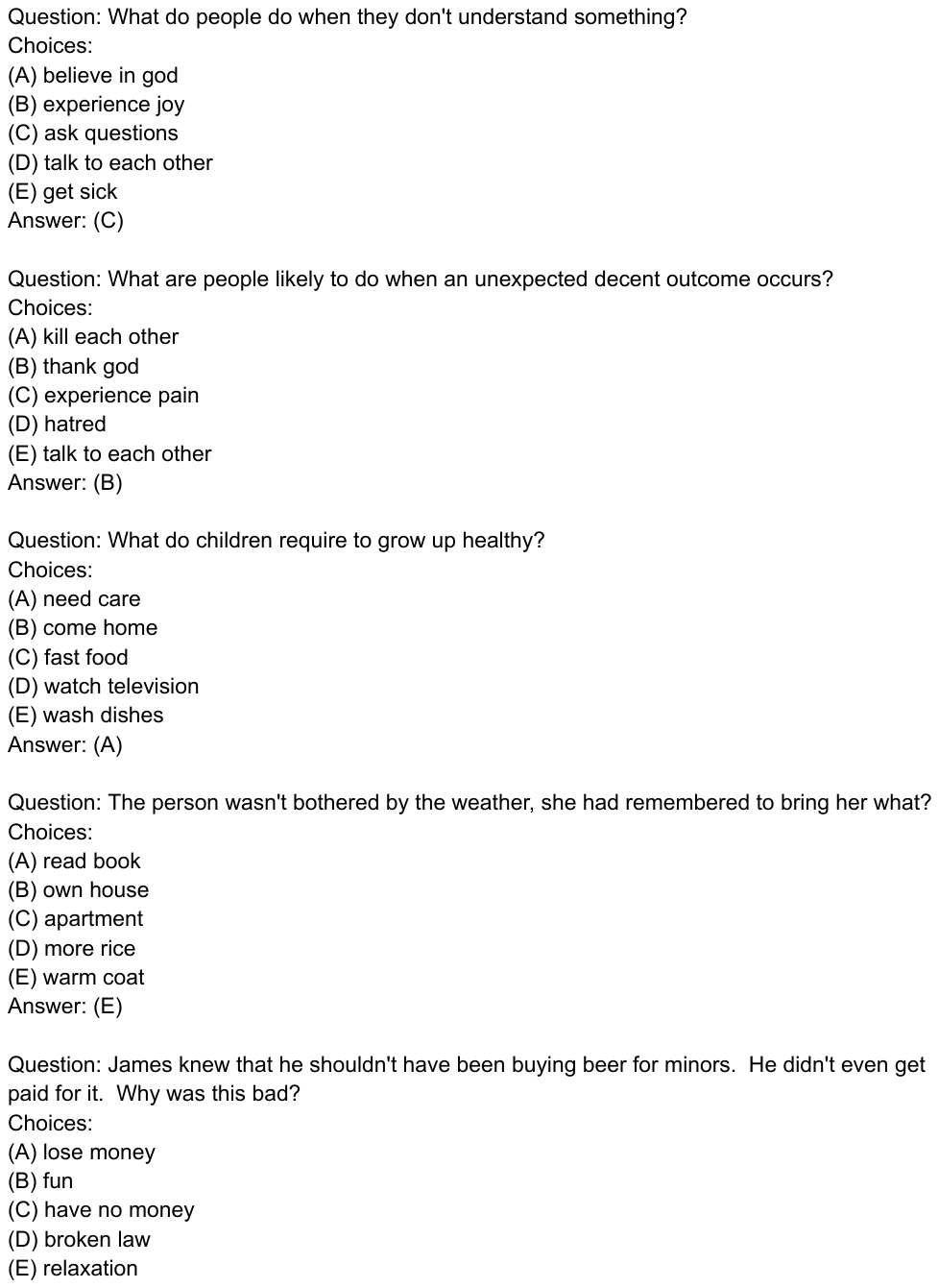}}
    \caption{In-context learning examples from CommonsenseQA for the full setting. (Part 1)}
\end{figure*}

\begin{figure*}
    \centering
    \fbox{\includegraphics[width=0.75\linewidth]{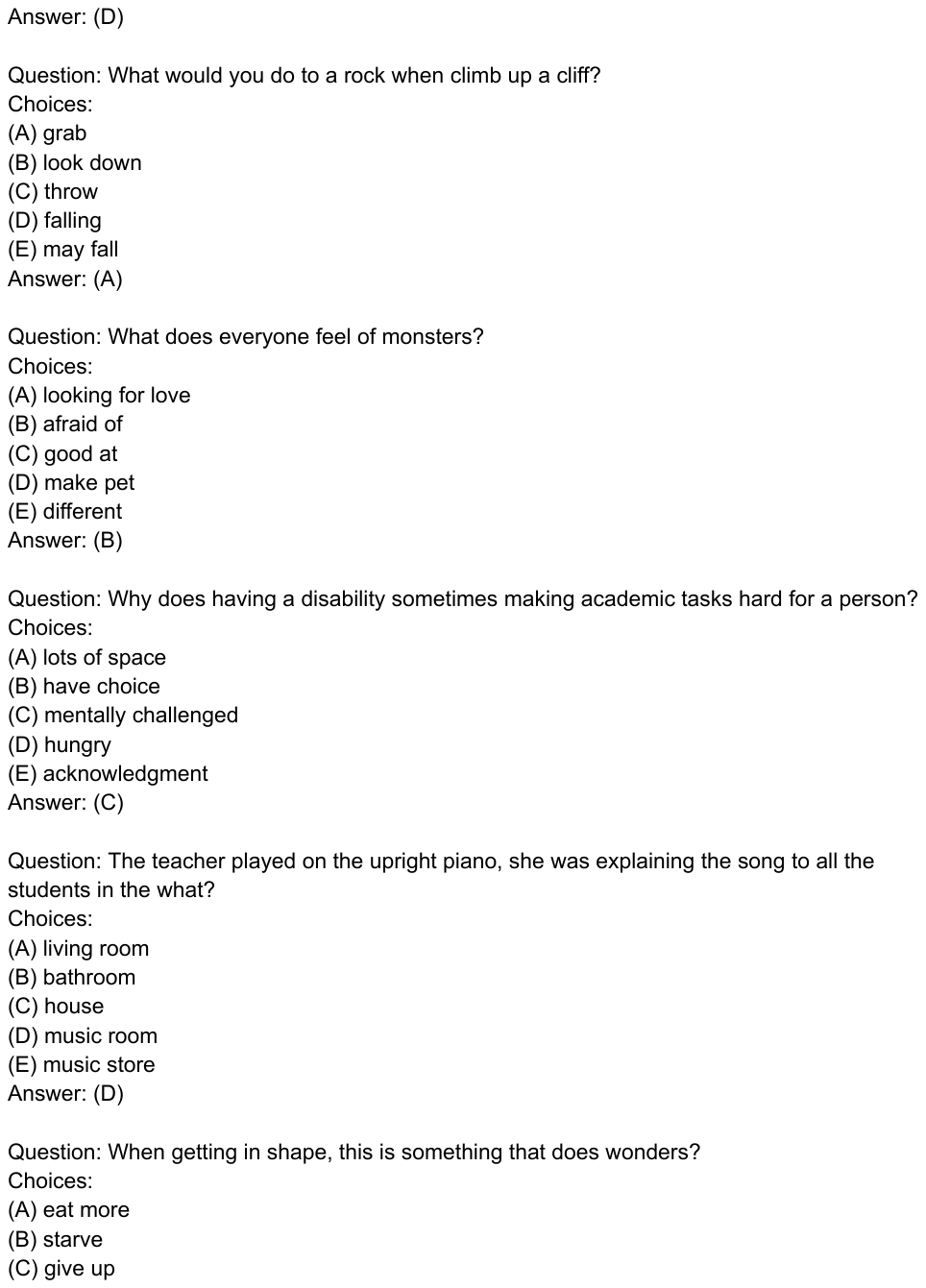}}
    \caption{In-context learning examples from CommonsenseQA for the full setting. (Part 2)}
\end{figure*}

\begin{figure*}
    \centering
    \fbox{\includegraphics[width=0.75\linewidth]{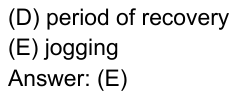}}
    \caption{In-context learning examples from CommonsenseQA for the full setting. (Part 3)}
\end{figure*}

%% file: data/full_ind_consistency.tex
\begin{table}[t]
\centering
\small
% \resizebox{0.9\columnwidth}{!}{
    \begin{tabular}{ l c c }
    \toprule
    \textbf{Agent}&\textbf{\abr{SIQA}} &  \textbf{\abr{CSQA}} \\
    \midrule
    LLaMA-2 7B & 38.3 & 22.7 \\
    LLaMA-2 13B & 53.3 & 56.4 \\
    LLaMA-2 70B & 60.1 & 58.9 \\
    Mistral 7B & 44.5 & 56.6 \\
    Mixtral 7x8B & 68.6 & 58.6 \\
    Yi 6B & 44.8 & 31.6 \\
    Yi 9B & 66.2 & 55.9 \\
    Yi 34B & 73.9 & 64.5 \\
    GPT-4 & 73.0 & 69.4 \\\midrule
    Average \abr{LLM} & 58.1 & 52.7 \\\midrule
    Human & 77.9 & 77.2 \\
    \bottomrule
    \end{tabular}
    % }
\caption{Percentage of cases where the most plausible answer from \cref{plausibility_ratings} matches the response to the full question from \cref{full_question_annotations}.}\label{tab:individual_to_full}
\end{table}

%% file: data/error.tex
\begin{table*}[ht]
\tiny
\centering
\begin{tabularx}{\linewidth}{c | X | X}
\toprule
\textbf{Label} & \multicolumn{1}{c|}{\textbf{Social IQA}} & \multicolumn{1}{c}{\textbf{CommonsenseQA}} \\ \midrule

\specialcell{Ambiguous}  & \specialcellleft{\textit{Context}: After seeing what a mess Aubrey was, Robin changed her into \\ clean clothes. \\
\textit{Question}: How would you describe Robin? \\
\textit{Choices}: \\
(A) \textcolor{red}{\textbf{a kind caretaker}} \\
(B) like a person who puts in thought \\
(C) \textcolor{blue}{\underline{a reliable friend}} \\ \textit{Explanation:} One needs to assume the relationship between \\ Aubrey and Robin to be able to pick a response.} & \specialcellleft{\textit{Question}: When you get together with friends to watch film, you might \\ do plenty of this? \\
\textit{Choices}: \\
(A) \textcolor{blue}{\underline{see what happens}} \\
(B) enjoy stories \\
(C) pass time \\
(D) \textcolor{red}{\textbf{have fun}} \\
(E) interesting \\ \textit{Explanation:} Answers B, C and D are all acceptable responses. \\ Answer A does not specify what one is actually ``seeing''.} \\ \midrule

\specialcell{Semantic Mismatch or Constraint} & \specialcellleft{\textit{Context}: Jesse just got a haircut and Riley was observing him  \\ with her eyes. \\
\textit{Question}: What will happen to Jesse? \\
\textit{Choices}: \\
(A) \textcolor{red}{\textbf{Give a compliment to Jesse about his hair}} \\
(B) \textcolor{blue}{\underline{go for a haircut}} \\
(C) see Jesse's haircut \\ \textit{Explanation:} None of the answer choices describe an event \\ that can ``happen'' to Jesse.} & \specialcellleft{\textit{Question}: What regions of a town would you have found a dime store? \\
\textit{Choices}: \\
(A) \textcolor{blue}{\underline{commercial building}} \\
(B) old movie \\
(C) \textcolor{red}{\textbf{small neighborhood}} \\
(D) past \\
(E) mall \\ \textit{Explanation:} Answers B and D are not ``regions of a town''.} \\ \midrule

\specialcell{Question is not coherent } & \specialcellleft{\textit{Context}: Remy answered the silly question they were asked happily. \\
\textit{Question}: Why did Remy do this? \\
\textit{Choices}: \\
(A) know the answer \\
(B) \textcolor{blue}{\underline{think about fun}} \\
(C) \textcolor{red}{\textbf{have fun}} \\ \textit{Explanation}: The question does not ask about anything \\ mentioned in the context. None of the answer choices are \\ a suitable response to the question.} & \specialcellleft{\textit{Question}: The flower grew tall to compete for sunlight, what did its  \\ neighbor do? \\
\textit{Choices}: \\
(A) \textcolor{blue}{\underline{blossom}} \\
(B) park \\
(C) open \\
(D) \textcolor{red}{\textbf{cast shadow}} \\
(E) vase \\ \textit{Explanation}: The question does not mention who ``neighbor'' \\ refers to.} \\ \midrule

\specialcell{No good answer choices} & \specialcellleft{\textit{Context}: Skylar wasn't certain that they had turned off the  \\ stove, so they went back to check. \\
\textit{Question}: What does Skylar need to do before this? \\
\textit{Choices}: \\
(A) \textcolor{red}{\textbf{anxious}} \\
(B) \textcolor{blue}{\underline{needed to have turned on the toaster}} \\
(C) good \\ \textit{Explanation}: None of the answer choices are \\ a suitable response to the question.} & \specialcellleft{\textit{Question}: What would a person need to do if his or her captain dies \\ at sea? \\
\textit{Choices}: \\ 
(A) cross street \\
(B) have a party \\
(C) \textcolor{blue}{\underline{experience life}} \\
(D) cross road \\
(E) \textcolor{red}{\textbf{man crew}} \\ \textit{Explanation}: None of the answer choices are a suitable response \\ to the question.} \\ \midrule

\specialcell{No prominent issue} & \specialcellleft{\textit{Context}: Robin had a hard time understanding the concept, so she let \\ Carson explain it more thoroughly. \\
\textit{Question}: How would Carson feel as a result? \\
\textit{Choices}: \\
(A) \textcolor{red}{\textbf{frustrated that Robin didn't understand}} \\
(B) ready to play \\
(C) \textcolor{blue}{\underline{ready to work}}} & \specialcellleft{\textit{Question}: What are people likely to do when an unexpected decent \\ outcome occurs? \\
\textit{Choices}: \\
(A) kill each other \\
(B) \textcolor{red}{\textbf{thank god}} \\
(C) experience pain \\
(D) hatred \\
(E) \textcolor{blue}{\underline{talk to each other}}} \\ \bottomrule

\end{tabularx}
\caption{\label{error_examples}Examples of ``plausibly problematic'' questions from \abr{SIQA} and \abr{CSQA} with labels. Text in \textcolor{blue}{blue} (also underlined) indicates \yplaus and text in \textcolor{red}{red} (also bolded) indicates \ydata. It is important to note that these labels are not mutually exclusive and a question can be ``plausibly problematic'' due to multiple reasons. Some of the above questions were tagged with more than one label, but we present unique questions for each label above.}
\end{table*}